\documentclass[10pt,twocolumn,letterpaper]{article}

\usepackage{cvpr}                

\usepackage{microtype}
\usepackage{multirow}
\usepackage{makecell}
\usepackage{placeins}
\usepackage{array}
\usepackage{mathtools}
\usepackage{amsthm}
\usepackage{algorithm}
\usepackage{algpseudocode}
\usepackage{colortbl}

\definecolor{sectionbg}{RGB}{255,246,236}
\definecolor{oursbg}{RGB}{224,244,252}
\definecolor{goodgreen}{RGB}{54,170,90}
\definecolor{badred}{RGB}{220,70,70}
\newcommand{\good}[1]{\textcolor{goodgreen}{\scriptsize (#1)}}
\newcommand{\bad}[1]{\textcolor{badred}{\scriptsize (#1)}}

\theoremstyle{plain}

\theoremstyle{definition}

\theoremstyle{remark}

\definecolor{cvprblue}{rgb}{0.21,0.49,0.74}
\usepackage[breaklinks,colorlinks,allcolors=cvprblue]{hyperref}

\def\paperID{*****} 
\def\confName{CVPR}
\def\confYear{2026}

\title{E2S-Pruner: Progressive Two-Stage Evidence Fusion for Visual Token Pruning in Vision-Language Models}

\author{Taoyu Qian$^{1}$ \quad Qi Wang$^{*1}$ \quad Daqian Shi$^{2}$ \quad
Yuanhao Jiang$^{3,4}$ \quad Shang Gao$^{1}$ \quad Hualong Yu$^{1}$\\[3pt]
{\small $^{1}$School of Computer Science and Engineering, Jiangsu University of Science and Technology}\\
{\small $^{2}$Digital Environment Research Institute (DERI), Queen Mary University of London}\\
{\small $^{3}$Shanghai Institute of Artificial Intelligence for Education, East China Normal University}\\
{\small $^{4}$National Institute of Education, Nanyang Technological University}\\
{\small $^{*}$Corresponding author: \texttt{wangqi@just.edu.cn}}}

\begin{document}
\maketitle

\begin{abstract}
Vision-language models typically encode an image into hundreds of visual tokens, incurring substantial inference latency and GPU memory overhead. Existing pruning methods largely rely on attention scores and directly aggregate outputs across attention heads and network layers, making it difficult to characterize evidential uncertainty and conflict. We propose E2S-Pruner, a progressive two-stage evidence-fusion framework for visual token pruning that requires no auxiliary model, trainable parameters, or fine-tuning. In the first stage, E2S-Pruner treats each attention head as an independent evidence source, estimates its reliability from evidence clarity and inter-head consistency, and represents each visual token using three states: important, unimportant, and uncertain. In the second stage, Dempster--Shafer evidence theory is used to quantify inter-layer conflict and fuse complementary evidence from multiple network layers. We further introduce a spatial novelty constraint that promotes coverage of distinct image regions and prevents the retained tokens from concentrating in a few locally salient areas. On LLaVA-1.5-7B, E2S-Pruner retains 98.0\%, 96.8\%, and 90.6\% of the aggregate performance when the average numbers of retained visual tokens are 192, 128, and 64, respectively, while improving throughput by $1.96\times$ and $2.09\times$ under the 128-token and 64-token settings. Experiments on Qwen2-VL-7B further demonstrate cross-model generalization. Code is available at \url{https://github.com/taoyu-qian/E2S-Pruner.git}.
\end{abstract}




\section{Introduction}
\begin{figure*}[!t]
\centering
\includegraphics[width=0.9\textwidth]{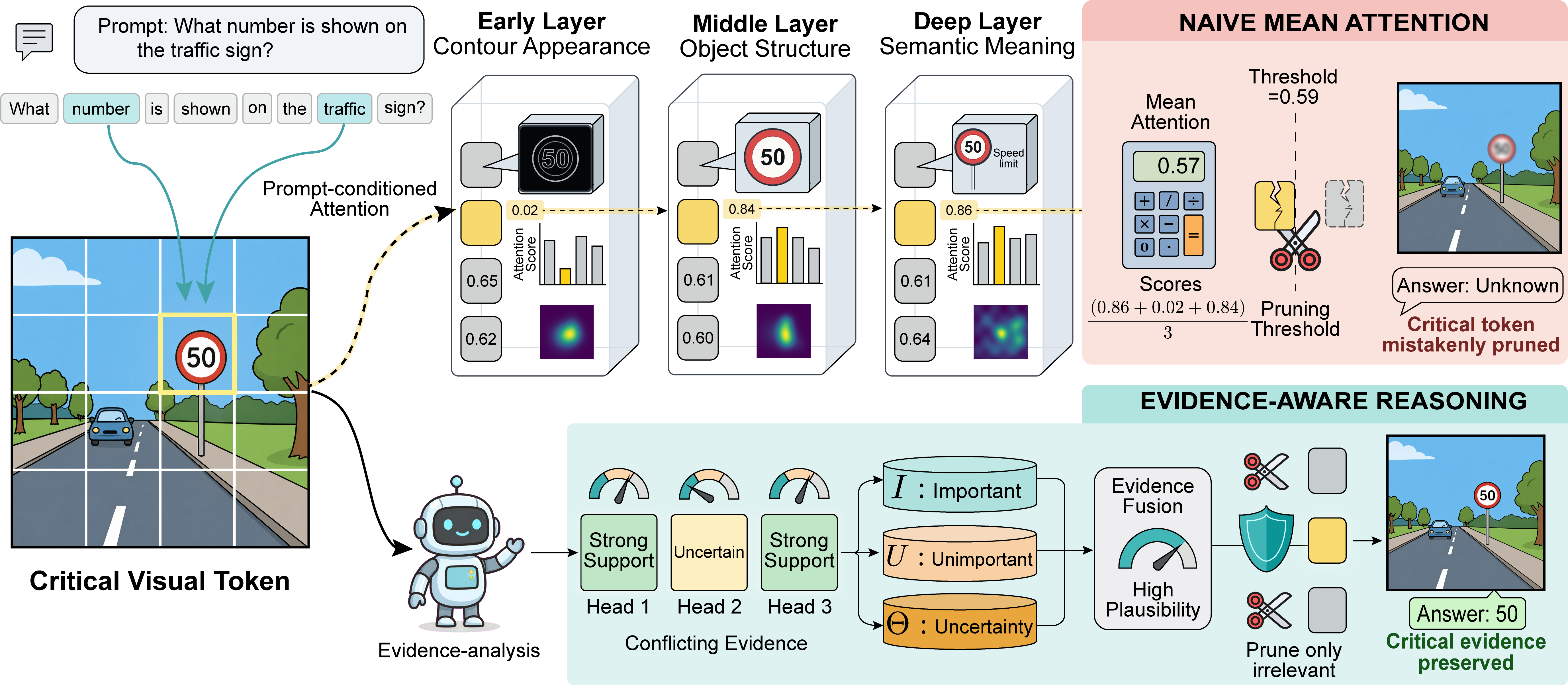}
\caption{Motivation and overview of E2S-Pruner. Direct aggregation can suppress tokens supported by complementary evidence across layers, whereas the proposed two-stage fusion explicitly models evidential uncertainty and conflict.}
\label{Fig:activate}
\end{figure*}

Recent advances in multimodal learning and large-scale pretraining have driven vision-language models (VLMs) from early exploration to rapid capability gains\cite{liu2024llava15,wang2024qwen2vl}. Meanwhile, edge-computing applications, including unmanned aerial vehicles\cite{wang2025jtduav} and autonomous driving systems\cite{pan2024vlp}, increasingly rely on the cross-modal understanding and reasoning capabilities of VLMs.

Covering the diversity and complexity of real-world tasks requires VLMs to learn from massive training corpora. Absorbing and representing such rich information generally entails a correspondingly large parameter count and substantial computation, which limits deployment in resource-constrained edge environments. This bottleneck has motivated a growing body of work on efficient VLM and visual Transformer architectures\cite{wang2024multitailed,su2024dctvit,jiang2025adfqvit}.

Most contemporary large language models (LLMs) are built on the Transformer architecture\cite{vaswani2017attention}, whose computational cost depends strongly on the input sequence length. Compressing redundant tokens can therefore reduce inference cost while preserving task performance. Existing token-compression methods fall broadly into two categories: (i) similarity-based token merging, represented by ToME\cite{bolya2023tome}, and (ii) attention-score-based token pruning, represented by FastV\cite{chen2024fastv}. The latter directly exploits attention information produced during inference and has consequently become a dominant approach. We adopt visual token pruning to improve VLM efficiency; its basic principle is introduced in Section~\ref{Section:Attention-Score-Based Pruning}.

We identify a limitation of existing approaches, illustrated in Fig.~\ref{Fig:activate}. Different attention heads and network layers exhibit distinct visual attention patterns\cite{cheng2025gradtoken}. Early LLM layers tend to capture low-level features such as local textures and edges, middle layers emphasize object-level features such as an individual road sign, and deep layers focus more strongly on global semantic structure. These representations therefore progress from local detail to high-level semantics\cite{chen2024fastv,arif2025hired}. To exploit this diversity, existing methods typically aggregate attention weights across heads and layers and retain the top-$K$ tokens according to the resulting scores. Consider the example at the top of Fig.~\ref{Fig:activate}, where three tokens receive scores of $(0.02,0.65,0.62)$, $(0.84,0.61,0.60)$, and $(0.86,0.61,0.64)$ in an early, middle, and deep layer, respectively. The first token is strongly supported by object-level and semantic evidence but lacks texture-level support. Simple averaging assigns it a score of only $0.57$, substantially weakening its apparent importance and increasing the risk of erroneous pruning. Thus, direct aggregation does not adequately model the joint distribution of evidence across heads and layers. A principled fusion mechanism that preserves complementary and conflicting information is required.

Unlike conventional probability theory, which assigns precise probabilities to individual hypotheses, Dempster--Shafer (D--S) evidence theory\cite{lefevre2002belief} supports belief assignment over sets of hypotheses and explicitly represents uncertainty during evidence fusion. It therefore provides a mathematical basis for modeling uncertain and conflicting information from multiple heads and layers. Building on this property, we introduce D--S evidence theory into visual token pruning and propose E2S-Pruner, a progressive two-stage evidence-fusion method. In the first stage, evidence from all heads within each selected early, middle, or deep LLM layer is fused to construct a D--S frame of discernment with three states: important, unimportant, and uncertain. In the second stage, inter-layer conflict is quantified and the layer-wise frames are recursively fused to obtain a global decision.

The main contributions of this work are summarized as follows:
\begin{itemize}
    \item We introduce D--S evidence theory into VLM token pruning. At the attention-head level, an evidence reliability measure is defined for each visual token to characterize agreement among heads, and the resulting reliability is used to construct a layer-wise D--S frame of discernment.
    
    \item We define an inter-layer conflict measure and use it to fuse evidence from multiple layer-wise frames. Belief and plausibility for visual-token importance are then derived within the D--S framework to determine token retention priority.
    
    \item We propose a spatial novelty constraint to prevent retained tokens from concentrating in a small image region. The image is partitioned into spatial cells, and a soft bonus is assigned to the highest-priority token in each cell to improve spatial coverage.
\end{itemize}

\section{Related Work}
\begin{figure}[t]
\centering
\includegraphics[width=\columnwidth]{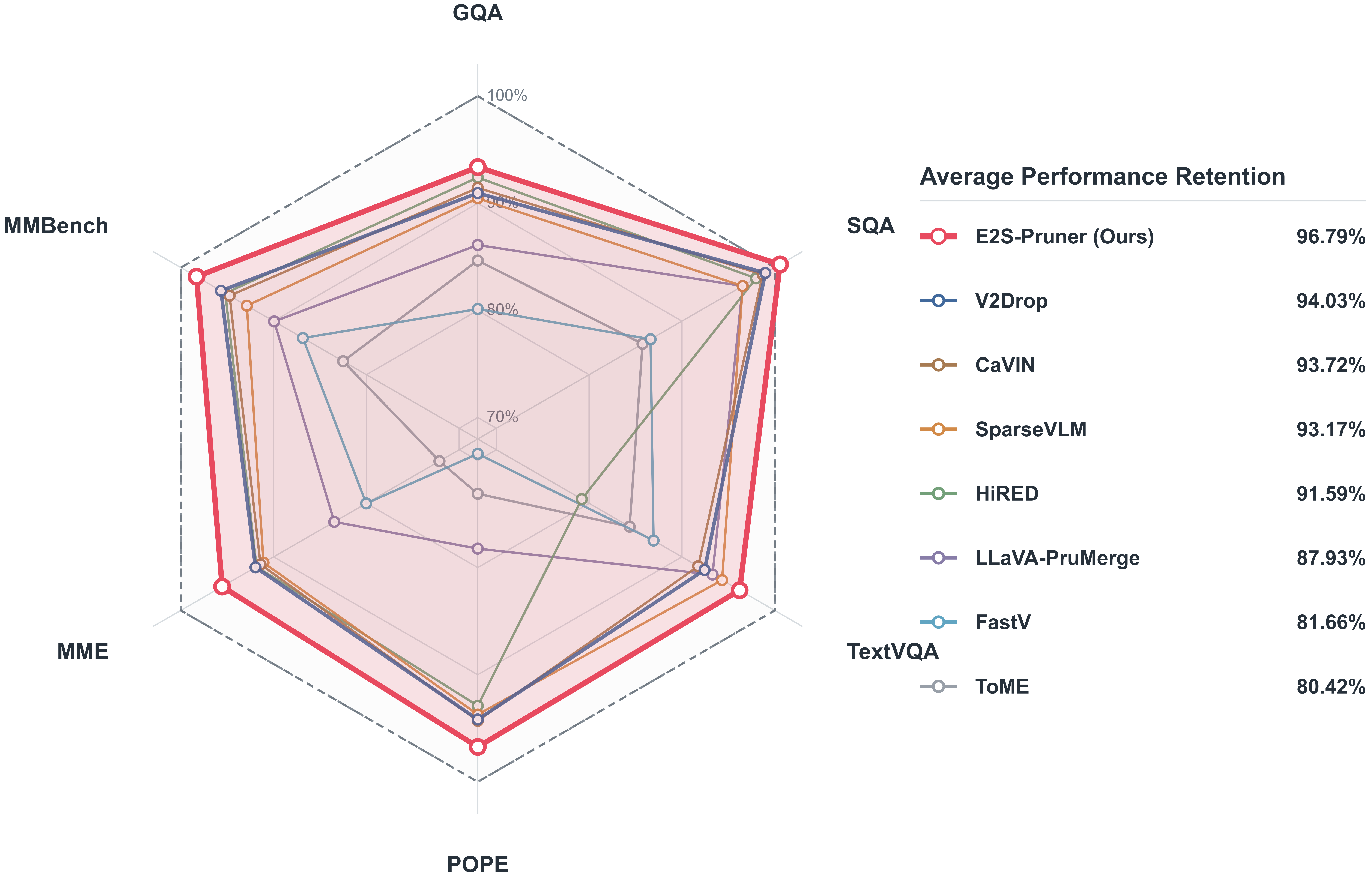}
\caption{Radar-chart comparison of E2S-Pruner on LLaVA-1.5-7B across the evaluated benchmarks, showing the balance between task performance and aggressive visual-token reduction.}
\label{Fig:rader}
\end{figure}

As image resolution and visual context length increase, visual tokens become a major source of VLM inference cost\cite{xian2025refining}. Existing visual token compression techniques can be broadly categorized as token merging or token pruning\cite{rao2021dynamicvit}. ToME\cite{bolya2023tome} progressively merges redundant tokens according to feature similarity and reduces Transformer computation without retraining. However, similarity-based merging lacks explicit guidance from the text instruction and may prematurely combine regions that are visually similar but semantically distinct. LLaVA-PruMerge\cite{shang2025prumerge} uses sparse class-to-patch attention in the vision encoder to identify key visual tokens and aggregates removed tokens into retained ones according to key-feature similarity, thereby mitigating information loss caused by direct pruning.

Another line of work estimates visual-token importance from internal attention or semantic relevance. FastV\cite{chen2024fastv} observes that deep LLM layers assign little attention to many visual tokens and therefore performs one-shot pruning in a shallow layer to reduce subsequent self-attention and feed-forward computation. HiRED\cite{arif2025hired} targets high-resolution inputs by using class-token attention in the vision encoder to allocate token budgets dynamically across image partitions and selecting informative tokens within each partition. SparseVLM\cite{zhang2025sparsevlm} identifies text tokens related to visual content, uses the LLM self-attention matrix to score visual tokens, and combines progressive sparsification, dynamic budget allocation, and token recycling. These methods establish the utility of attention for visual-token selection, but they generally aggregate scores directly across heads or layers without explicitly modeling source reliability, conflict, or uncertainty.

Beyond saliency scoring, several studies reconsider visual token compression from the perspectives of redundancy, representation dynamics, and causal association. DART\cite{wen2025dart} argues that token importance alone is not a sufficiently reliable pruning criterion and retains complementary information by measuring redundancy between candidate tokens and a small set of pivot tokens. V2Drop\cite{chen2026v2drop} progressively prunes tokens according to representation changes between adjacent LLM layers, preferentially removing slowly changing tokens to reduce positional bias while remaining compatible with efficient attention kernels. CaVIN\cite{qian2026cavin} further notes that conventional relevance scores may overlook dependencies between target regions and their context; it therefore learns visual-token associations from weak causal signals in chain-of-thought data and recovers contextual tokens that conventional pruning may discard.

Although these approaches improve compression through attention saliency, feature similarity, information redundancy, inter-layer dynamics, or contextual dependence, evidence quality varies across attention heads and network layers. Simple summation or averaging can obscure local details, complementary semantics, and potentially important visual evidence associated with high conflict. E2S-Pruner addresses this limitation by jointly modeling multi-head and multi-layer information with D--S evidence theory and by improving regional coverage through a spatial constraint, yielding a more stable performance--efficiency trade-off across token budgets. Figure~\ref{Fig:rader} summarizes its performance on LLaVA-1.5-7B across multiple benchmarks.

\FloatBarrier

\section{Preliminary}
\subsection{Attention-Score-Based Pruning} \label{Section:Attention-Score-Based Pruning}
A VLM comprises a vision encoder and an LLM. Images and text are encoded as visual and text tokens, respectively, before being passed to the LLM. Because visual tokens substantially outnumber text tokens, pruning is applied primarily to the visual sequence. Attention-based pruning has several variants; in this work, visual-token importance is measured using text-to-visual attention. This head-level representation exploits the interaction between the attention mechanism and the CLIP vision encoder\cite{radford2021clip}: intermediate visual tokens encode rich image features while incorporating semantic guidance from the textual prompt, providing discriminative evidence for token selection.

For the $h$-th head in the $l$-th LLM layer, let $N_T$ and $N_V$ denote the numbers of text and visual tokens, respectively, where $N_V \gg N_T$. The two token sets are $\mathcal{T}=\{T_1,T_2,\ldots,T_{N_T}\}$ and $\mathcal{V}=\{V_1,V_2,\ldots,V_{N_V}\}$. The attention score of the $n$-th visual token in this head is
\begin{equation}
a_{l,h,n}  = \frac{1}{N_T}\sum_{t \in \mathcal{T}}\frac{Q_t K_{v_n}^T}{\sqrt{d}}
\end{equation}

where $Q_t$ is the query of the $t$-th text token, $K_{v_n}$ is the key of the $n$-th visual token, and $d$ is the head dimension.

Conventional pruning methods first aggregate the single-head scores $a_{l,h,n}$ across heads and then across layers to obtain one attention-based importance score for each visual token. Tokens are ranked by this score, and the top $K$ are retained.

\subsection{D-S Theory}
For a decision problem, the set of all possible hypotheses is called the frame of discernment and is denoted by $\Theta$. Evidence comprises observations, computed quantities, or prior knowledge that support, oppose, or remain uncertain about propositions in this frame. Different information sources may yield distinct assessments of the same problem; $n$ such sources form a collection $\mathcal{E}=\{\Theta_1,\Theta_2,\ldots,\Theta_n\}$. D--S evidence theory provides a principled mechanism for fusing these sources.

For any subset $A\subseteq\Theta$, the basic probability assignment $m(A)$ quantifies the support assigned to proposition $A$ and satisfies
\begin{equation}
\begin{aligned}
m&(\emptyset) = 0 \\
\sum_{A\subset\Theta}&m(A) = 1
\end{aligned}
\end{equation}

Given propositions $B$ and $C$ from two evidence sources, their conflict degree is defined as
\begin{equation}
K = \sum_{B\cap C = \emptyset}m_1(B)m_2(C) \qquad K\in[0,1]
\end{equation}

This expression sums the products of masses assigned to mutually exclusive proposition pairs and thereby quantifies global conflict between the two sources. A value of $K$ closer to 1 indicates stronger disagreement.

Dempster's rule combines the two sources into the following mass function:
\begin{equation}
m(A) = \frac{\sum\limits_{B\cap C=A}m_1(B)m_2(C)}{1-K}, \quad A\neq \emptyset
\end{equation}

Unlike conventional probability theory, D--S theory allows mass to be assigned directly to a composite proposition. For example,
\begin{equation}
m(\{A,B \}) = 0.4
\end{equation}

indicates that the evidence supports ``the outcome is $A$ or $B$'' but cannot distinguish between the two. Similarly, $m(\Theta)$ represents mass assigned to complete uncertainty.

Given a mass function, the belief function accumulates all masses that fully support $A$ and therefore gives a lower bound on its support:
\begin{equation}
\operatorname{Bel}(A) = \sum_{B \subseteq A}m(B)
\end{equation}

The plausibility function accumulates all masses that do not contradict $A$ and therefore gives an upper bound on its support:
\begin{equation}
\operatorname{Pl}(A) = \sum_{B \cap A \neq \emptyset}m(B)
\end{equation}

\section{Methodology}
\begin{algorithm}[!t]
\caption{E2S-Pruner}
\label{ALG:E2S}
\begin{algorithmic}
\Require Text tokens $\mathcal{T}$; visual tokens $\mathcal{V}$;
pruning layers $\mathcal{P}$; budgets $\{K_l^{\mathrm{keep}}\}$;
$G=4$ and $\lambda=0.1$
\Ensure Retained visual tokens $\mathcal{V}$

\For{$l\in\mathcal{P}$}
    \ForAll{$v_n\in\mathcal{V}$ and attention heads $h$}
        \State Compute $a_{l,h,n}$ using text-to-visual attention
        \State Normalize:
        $r_{l,h,n}\gets\operatorname{Softmax}_{n}(a_{l,h,n})$
        \State Map to support:
        $p_{l,h,n}\gets
        \frac{N_l r_{l,h,n}}{1+N_l r_{l,h,n}}$
    \EndFor

    \ForAll{$v_n\in\mathcal{V}$}
        \State Compute reliability $\rho_{l,n}$ from head clarity
        and consistency
        \State Construct
        \[
        m_{l,n}(I)=\rho_{l,n}\bar p_{l,n},\quad
        m_{l,n}(U)=\rho_{l,n}(1-\bar p_{l,n}),\quad
        m_{l,n}(\Theta)=1-\rho_{l,n}
        \]
        \State Recursively fuse $m_{l,n}$ with previous-layer evidence
        using Dempster's rule to obtain $\widetilde m_{n}^{(l)}$
        \State Compute
        \[
        q_{l,n}\gets
        1-\bigl(1-\operatorname{Pl}_{n}^{(l)}(I)\bigr)
        \bigl(1-K_{n}^{(l)}\bigr)
        \]
    \EndFor

    \State Select the highest-$q$ representative in each occupied
    $G\times G$ spatial cell
    \State $\widetilde q_{l,n}\gets
    (1-\lambda)q_{l,n}+\lambda b_{l,n}$
    \State $\mathcal{V}\gets
    \operatorname{TopK}
    (\{\widetilde q_{l,n}\},K_l^{\mathrm{keep}})$
\EndFor
\State \Return $\mathcal{V}$
\end{algorithmic}
\end{algorithm}

This section presents the E2S-Pruner algorithm. Its overall workflow is illustrated in Fig.~\ref{Fig:activate}, and the pseudocode is provided in Algorithm~\ref{ALG:E2S}. Three pruning operations are deployed in shallow, middle, and deep LLM layers. The visual-token count decreases stepwise as pruning progresses, producing a coarse-to-fine compression schedule.

\subsection{Attention-head Fusion}
\begin{figure*}[!t]
\centering
\includegraphics[width=0.84\textwidth]{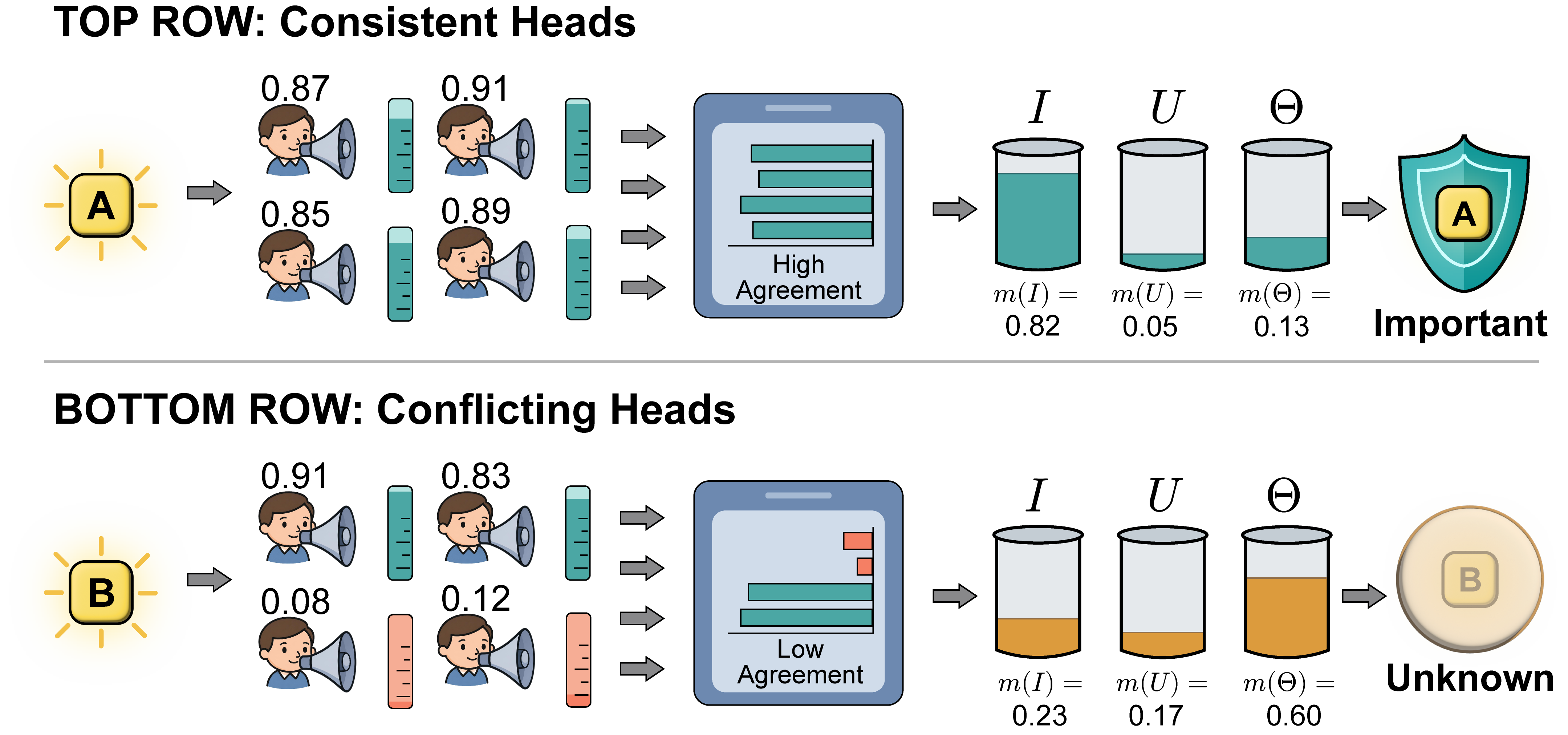}
\caption{Reliability-aware fusion of multi-head evidence. Clear and consistent head responses receive high reliability, whereas ambiguous or conflicting responses contribute greater uncertainty.}
\label{Fig:Vote}
\end{figure*}

To address inconsistent attention responses across heads and layers, we use D--S evidence theory to model their conflict explicitly. This subsection introduces the first stage: multi-head evidence fusion.

Each selected LLM layer yields a frame of discernment $\Theta$ for every candidate token. The frames from $k$ pruning layers form the collection $\mathcal{E}=\{\Theta_1,\Theta_2,\ldots,\Theta_k\}$. Each layer-wise frame is constructed as follows.

Consider the $h$-th attention head in the $l$-th pruning layer, where $l$ indexes only layers at which pruning is performed. Equation~(1) gives the attention score of the $n$-th visual token $v_n$ in this head. We normalize the scores as
\begin{equation}
r_{l,h,n} = \frac{\operatorname{exp}(a_{l,h,n})}{\sum_{j}\operatorname{exp}(a_{l,h,j})}
\end{equation}

We then define the support score
\begin{equation}
\label{Eq:support}
p_{l,h,n} = \frac{N_l \cdot r_{l,h,n}}{1+N_l \cdot r_{l,h,n}}
\end{equation}

This mapping converts relative attention into a support score in $[0,1)$. Because the mean normalized attention over the $N_l$ tokens in pruning layer $l$ is $1/N_l$, $r_{l,h,n}=1/N_l$ gives $p_{l,h,n}=0.5$. This value corresponds to average attention and indicates no clear selection preference.

As illustrated in Fig.~\ref{Fig:Vote}, suppose that the current layer has four attention heads. For a given token, their support scores may be $[0.87,0.91,0.85,0.89]$, indicating consistent support and reliable evidence. Alternatively, the scores may be $[0.91,0.83,0.08,0.12]$, indicating severe inter-head conflict. Direct averaging would obscure this distinction. We therefore compute the evidence reliability $\rho_{l,n}$ as
\begin{equation}
\rho_{l,n} = \frac{2}{H}\bigg|\sum_{h=1}^H(p_{l,h,n}-0.5)\bigg| \cdot \bigg(1-\frac{2}{H}\sum_{h=1}^H|p_{l,h,n}-\bar p_{l,n}| \bigg)
\end{equation}

Evidence reliability is defined as the product of decision clarity and support consistency. The first factor measures the selection clarity of individual heads, whereas the second characterizes inter-head agreement through deviations from the mean score.

D--S theory represents three states: (i) important, denoted by $I$, when the evidence supports retaining $v_n$; (ii) unimportant, denoted by $U$, when the evidence does not support retaining $v_n$; and (iii) uncertain, denoted by $\Theta$, when the evidence is insufficient or conflicting. The frame is therefore $\Theta=\{I,U\}$. The mass function of the $n$-th candidate token in pruning layer $l$ is
\begin{equation}
\begin{aligned}
m_{l,n}(I) &= \rho_{l,n} \bar p_{l,n} \\
m_{l,n}(U) &= \rho_{l,n} (1- \bar p_{l,n}) \\
m_{l,n}(\Theta) &= 1 - \rho_{l,n}
\end{aligned}
\end{equation}

The model now represents each token not by a single attention magnitude but by mass assigned to the important, unimportant, and uncertain states. Disagreement among attention heads is thus quantified as uncertainty within the frame of discernment. This representation avoids premature decisions that could remove critical information and provides a discriminative basis for subsequent multi-source fusion.

\subsection{Layer-wise Evidence Fusion}
To balance inference efficiency and evidential completeness, multi-layer fusion is performed only at the designated pruning layers. This design exploits edge and texture cues from shallow layers, object structure from middle layers, and global semantics from deep layers. Evidence is recursively fused in pruning-layer order. Let $\widetilde{m}_{n}^{(l)}$ denote the cumulative mass function of token $n$ after fusion through pruning layer $l$. It is initialized as
\begin{equation}
\widetilde{m}_{n}^{(1)}(A)=m_{1,n}(A),
\qquad A\in\{I,U,\Theta\}
\end{equation}

At pruning layer $l$, we first compute the conflict between $m_{l,n}$ and the accumulated historical mass $\widetilde{m}_{n}^{(l-1)}$:
\begin{equation}
K_{n}^{(l)}
=
\widetilde{m}_{n}^{(l-1)}(I)m_{l,n}(U)
+
\widetilde{m}_{n}^{(l-1)}(U)m_{l,n}(I)
\end{equation}

The fused mass assigned to the important hypothesis $I$ is
\begin{equation}
\begin{aligned}
\widetilde{m}_{n}^{(l)}(I)
=
\frac{1}{1-K_{n}^{(l)}}
\Bigl[
\widetilde{m}_{n}^{(l-1)}(I)m_{l,n}(I)
+
\widetilde{m}_{n}^{(l-1)}(I)m_{l,n}(\Theta)\\
+
\widetilde{m}_{n}^{(l-1)}(\Theta)m_{l,n}(I)
\Bigr]
\end{aligned}
\end{equation}

The fused mass assigned to the unimportant hypothesis $U$ is
\begin{equation}
\begin{aligned}
\widetilde{m}_{n}^{(l)}(U)
=
\frac{1}{1-K_{n}^{(l)}}
\Bigl[
\widetilde{m}_{n}^{(l-1)}(U)m_{l,n}(U)
+
\widetilde{m}_{n}^{(l-1)}(U)m_{l,n}(\Theta)\\
+\widetilde{m}_{n}^{(l-1)}(\Theta)m_{l,n}(U)
\Bigr]
\end{aligned}
\end{equation}

The fused mass assigned to uncertainty $\Theta$ is
\begin{equation}
\widetilde{m}_{n}^{(l)}(\Theta)
=
\frac{
\widetilde{m}_{n}^{(l-1)}(\Theta)m_{l,n}(\Theta)
}{
1-K_{n}^{(l)}
}
\end{equation}

This yields the fused mass function $\widetilde{m}_{n}^{(l)}$ for token $n$ at pruning layer $l$. Its belief and plausibility for importance hypothesis $I$ are, respectively,
\begin{equation}
\operatorname{Bel}_{n}^{(l)}(I)
=
\widetilde{m}_{n}^{(l)}(I)
\end{equation}
\begin{equation}
\operatorname{Pl}_{n}^{(l)}(I)
=
\widetilde{m}_{n}^{(l)}(I)
+
\widetilde{m}_{n}^{(l)}(\Theta)
=
1-\widetilde{m}_{n}^{(l)}(U)
\end{equation}

To avoid incorrectly pruning tokens with insufficient evidence, we use plausibility rather than belief as the pruning criterion. The retention priority at pruning layer $l$ is defined as
\begin{equation}
q_{l,n}= 1-\big(1-Pl_n^{(l)}(I) \big)(1-K_n^{(l)})
\end{equation}
where $q_n\in[0,1]$ is the token retention-priority score. Tokens with greater plausibility or stronger conflict receive higher scores.

\subsection{Spatial Novelty Regularization and Token Pruning}

\begin{figure}[!t]
\centering
\includegraphics[width=\columnwidth]{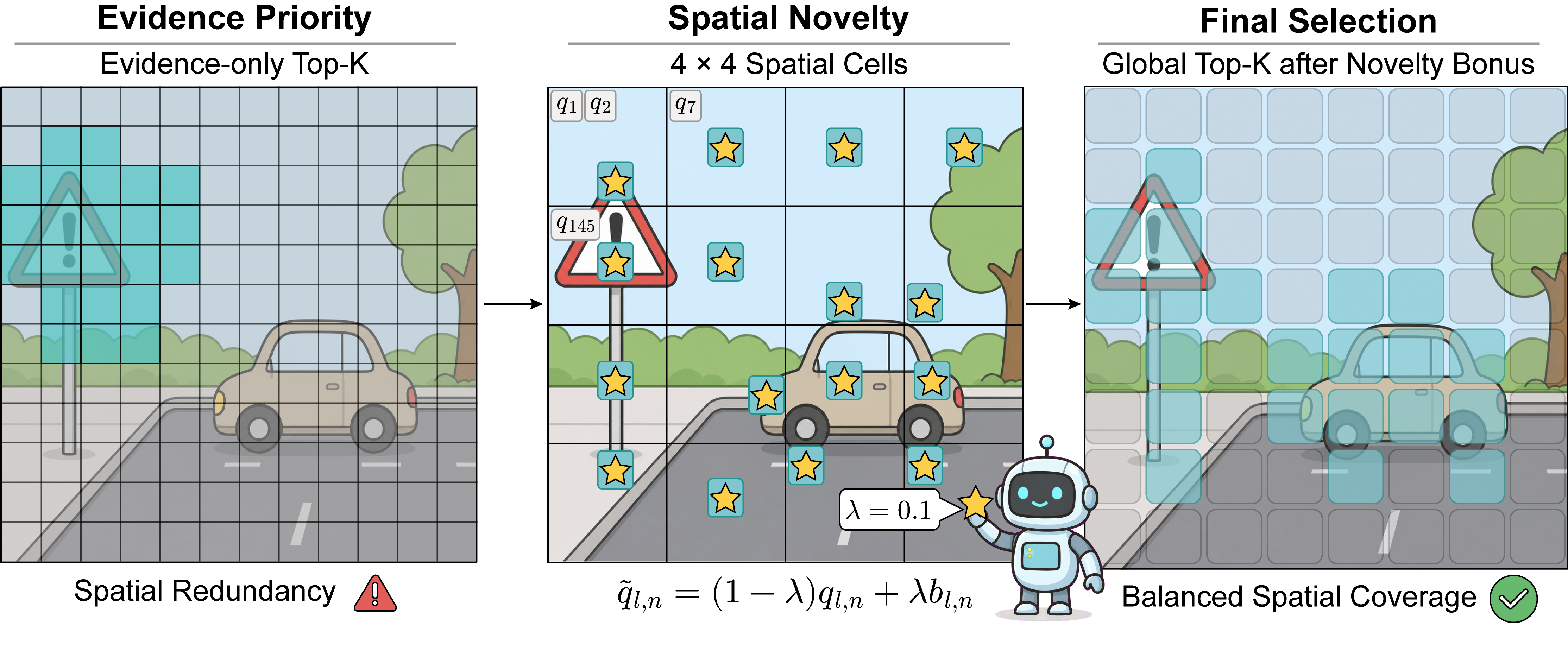}
\caption{Spatial novelty regularization. The visual-token grid is partitioned into spatial cells, and the highest-priority candidate in each occupied cell receives a soft bonus before global Top-$K$ selection.}
\label{Fig:Space}
\end{figure}

Although the evidence-based retention priority reflects semantic relevance between visual tokens and the text query, direct global Top-$K$ selection may concentrate the retained tokens in a small image region and discard useful information elsewhere. We therefore introduce the lightweight spatial novelty constraint shown in Fig.~\ref{Fig:Space}. It improves the spatial coverage of retained visual tokens without altering the evidence-fusion result.

For LLaVA-1.5-7B, the initial 576 visual tokens form a $24\times24$ grid. We partition this grid into $G\times G$ spatial cells. With $G=4$, each cell contains $6\times6$ visual tokens. Let $\mathcal{V}_l$ denote the candidate set at the current layer and $\mathcal{V}_{l,g}$ the candidates in spatial cell $g$. The representative token of that cell is
\begin{equation}
r_{l,g}=\arg\max_{v_n\in\mathcal{V}_{l,g}} q_{l,n}
\end{equation}

This operation identifies the highest-priority token in each spatial cell. We further define the spatial representative indicator
\begin{equation}
b_{l,n}=
\begin{cases}
1, & v_n=r_{l,g}\text{ for some occupied cell }g,\\
0, & \text{otherwise}.
\end{cases}
\end{equation}

The spatially adjusted score is
\begin{equation}
\widetilde{q}_{l,n}
=
(1-\lambda)q_{l,n}+\lambda b_{l,n}
\end{equation}
where $\lambda$ is the spatial novelty weight, set to $0.1$ in all experiments. Global Top-$K$ selection is then performed according to $\widetilde{q}_n$:
\begin{equation}
\mathcal{V}_{\mathrm{keep}}
=
\operatorname{TopK}\left(
\{\widetilde{q}_{l,n}\mid v_n\in\mathcal{V}\},K
\right)
\end{equation}

where $\mathcal{V}_{\mathrm{keep}}$ is the set of visual tokens retained after pruning the current layer.

This mechanism applies only a soft bonus to the representative token with the highest evidence score in each occupied cell; it does not force every cell to retain a token. Consequently, it preserves the global evidence-based ranking while reducing excessive concentration of visual tokens in local regions.

\FloatBarrier

\section{Experiments and Analysis}

All experiments were conducted on an NVIDIA GeForce RTX~4080 Super GPU with 32~GB of memory. This section presents the experimental protocol and analyzes the results of E2S-Pruner.

\subsection{Experiment Settings}
\subsubsection{Datasets}
We use seven benchmarks to evaluate pruning performance from complementary perspectives. The main experiments adopt LLaVA-1.5-7B\cite{liu2024llava15} as the dense baseline and evaluate on GQA, SQA, TextVQA, POPE, MME, and MMBench. Cross-model experiments use Qwen2-VL-7B\cite{wang2024qwen2vl} and evaluate on SQA, POPE, MME, and AI2D. For fair and reproducible comparison, all datasets use their official prompt templates and evaluation code.

\begin{itemize}
    \item \textbf{GQA}\cite{hudson2019gqa} is a comprehensive visual question answering benchmark built from real images and structured scene representations. It emphasizes reasoning over object attributes, spatial relations, and compositional semantics rather than simple object recognition.
    \item \textbf{ScienceQA (SQA)}\cite{lu2022scienceqa} contains multiple-choice science questions from elementary- and secondary-school curricula. Some questions include images, charts, or diagrams, enabling evaluation of textual understanding, visual comprehension, and scientific reasoning.
    \item \textbf{TextVQA}\cite{singh2019textvqa} evaluates the ability to read and reason about text in images. Answering its questions often requires recognizing scene text, menus, signs, or document content and integrating it with visual context.
    \item \textbf{POPE}\cite{li2023pope}, or Polling-based Object Probing Evaluation, measures object hallucination in VLMs by asking whether specific objects are present and testing whether the model incorrectly reports nonexistent objects.
    \item \textbf{MME}\cite{fu2025mme} is a comprehensive multimodal benchmark spanning visual perception and cognitive reasoning. Its tasks assess object existence, counting, position, optical character recognition, commonsense knowledge, and reasoning.
    \item \textbf{MMBench}\cite{liu2024mmbench} is a multiple-choice benchmark for multimodal large models that covers fine-grained abilities such as attribute recognition, spatial relations, logical reasoning, and chart understanding.
    \item \textbf{AI2D}\cite{kembhavi2016ai2d} evaluates the understanding of scientific diagrams and instructional illustrations. It requires models to parse objects, structures, and relations and to reason jointly over visual and textual information.
\end{itemize}

\subsubsection{Pruning Settings}
Although pruning methods differ in pruning locations and the number of tokens removed at each operation, they are compared under the same average-token budget. The average number of retained tokens is defined as
\begin{equation}
\mathrm{AvgTokens} = \frac{1}{N}\sum_{i=1}^N T_i
\end{equation}

where $N$ is the number of LLM layers and $T_i$ is the number of visual tokens retained in layer $i$.

Existing pruning schedules can be grouped into three types: (i) reducing the sequence to $\mathrm{AvgTokens}$ before it enters the LLM; (ii) performing one pruning operation at an intermediate LLM layer; and (iii) pruning progressively at multiple intermediate layers. We adopt the third strategy and evaluate average-token budgets of 192, 128, and 64.

LLaVA-1.5-7B has 32 LLM layers and receives 576 visual tokens. We perform three pruning operations after layers 3, 17, and 22; consequently, the visual-token count decreases at the inputs to layers 4, 18, and 23. For average-token budgets of 192, 128, and 64, the three operations retain $194\rightarrow148\rightarrow96$, $122\rightarrow48\rightarrow42$, and $20\rightarrow8\rightarrow0$ visual tokens, respectively.

\begin{figure*}[!t]
\centering
\includegraphics[width=0.90\textwidth]{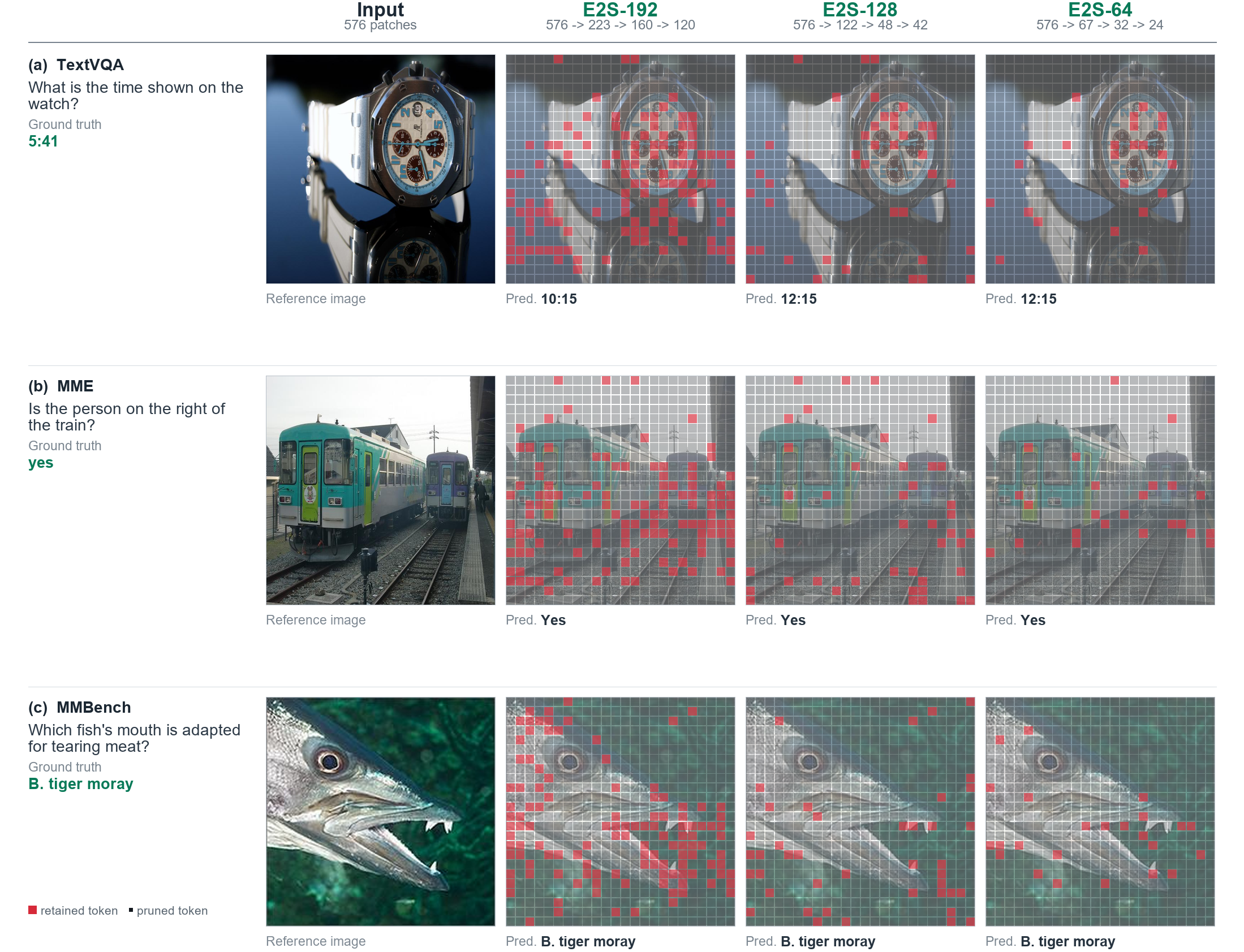}
\caption{Qualitative visualization of final-layer visual-token retention under average budgets of 192, 128, and 64 tokens across representative TextVQA, ScienceQA, and MMBench examples. Green markers denote retained tokens, whereas red points indicate pruned patch locations.}
\label{Fig:Show1}
\end{figure*}

Figure~\ref{Fig:Show1} visualizes the pruning outcomes for the same examples under different average-token budgets. Green markers indicate the visual tokens retained in the final layer.

\subsubsection{Evaluation Metrics}
Performance on every dataset is measured using its official evaluation protocol. To summarize performance across heterogeneous benchmarks, we compute a normalized performance-retention score relative to the dense baseline. For LLaVA-1.5-7B, this score is
\begin{equation}
\begin{aligned}
\mathrm{Retention}
=
\frac{1}{6}
\Bigg(
&\frac{S_{\mathrm{GQA}}}{61.9}
+\frac{S_{\mathrm{SQA}}}{69.5}
+\frac{S_{\mathrm{TextVQA}}}{58.2}
+\frac{S_{\mathrm{POPE}}}{85.9}
\\
&+\frac{S_{\mathrm{MME}}}{1862}
+\frac{S_{\mathrm{MMBench}}}{64.6}
\Bigg)
\times 100\%.
\end{aligned}
\label{eq:performance_retention}
\end{equation}

where each numerator is the score of the pruned model on one benchmark and the corresponding denominator is the score of the dense baseline.

\subsection{Comparative Experiment}
Table~\ref{tab:main_comparison} compares E2S-Pruner with representative visual-token pruning methods under three average-token budgets. E2S-Pruner achieves the highest aggregate performance at every budget. With average budgets of 192, 128, and 64 visual tokens, it retains 98.0\%, 96.8\%, and 90.6\% of the dense LLaVA-1.5-7B performance, respectively, demonstrating that it preserves multimodal understanding despite substantial token reduction.

At an average budget of 192 tokens, E2S-Pruner retains 98.0\% of aggregate performance, exceeding SparseVLM and V2Drop by 2.1 and 0.4 percentage points, respectively. It obtains the best results on SQA, TextVQA, and MMBench, outperforming V2Drop by 0.5, 1.6, and 0.7 points on these benchmarks. Although V2Drop is slightly better on POPE and MME, the higher aggregate retention of E2S-Pruner indicates a more balanced preservation of capabilities across multimodal tasks.

The advantage becomes more pronounced at an average budget of 128 tokens. E2S-Pruner achieves the best result among the compared methods on all six benchmarks and retains 96.8\% of aggregate performance. Relative to V2Drop, it improves GQA, SQA, TextVQA, POPE, MME, and MMBench by 1.5, 1.1, 2.2, 2.2, 67, and 1.7 points, respectively, and improves aggregate retention by 2.8 percentage points. It also exceeds SparseVLM by 3.6 percentage points in aggregate retention. These results suggest that E2S-Pruner identifies and preserves task-relevant visual evidence more accurately than methods driven primarily by local attention magnitude or token redundancy.

Under the more challenging 64-token budget, E2S-Pruner still retains 90.6\% of dense-model performance, exceeding V2Drop and SparseVLM by 3.7 and 4.1 percentage points, respectively. Compared with V2Drop, the gains on GQA, SQA, POPE, MME, and MMBench are 5.0, 0.5, 3.4, 186, and 2.1 points. The particularly large gains on GQA and MME show that E2S-Pruner preserves visual information essential to image understanding and multimodal reasoning even after most visual tokens are removed.

One limitation emerges on TextVQA under the 64-token setting: E2S-Pruner scores 49.5, below the 54.0 achieved by LLaVA-PruMerge. At extremely low budgets, aggressive pruning may remove image regions containing fine-grained text and therefore disproportionately affect scene-text recognition and reasoning. Protecting textual regions under severe compression is an important direction for future work.

Across the three budgets, E2S-Pruner degrades more slowly than most competing methods as the pruning rate increases. Aggregate retention decreases from 98.0\% at 192 tokens to 96.8\% at 128 tokens and remains 90.6\% at 64 tokens. This trend indicates that the proposed components jointly improve the accuracy and stability of visual-token importance estimation, making E2S-Pruner particularly suitable for resource-constrained applications or scenarios requiring aggressive compression.

\begin{table*}[t]
    \centering
\caption{Performance comparison of visual-token pruning methods on LLaVA-1.5-7B, with average scores normalized to the dense baseline.}
    \label{tab:main_comparison}

    \setlength{\tabcolsep}{5.5pt}
    \renewcommand{\arraystretch}{1.12}

    \resizebox{\textwidth}{!}{%
    \begin{tabular}{lccccccc}
        \toprule
        \textbf{Method}
        & \textbf{GQA}
        & \textbf{SQA}
        & \textbf{TextVQA}
        & \textbf{POPE}
        & \textbf{MME}
        & \textbf{MMBench}
        & \textbf{Avg} \\
        \midrule

        \rowcolor{sectionbg}
        \multicolumn{8}{c}{\itshape Vanilla 576 Tokens} \\

        LLaVA-1.5-7B
        & 61.9 & 69.5 & 58.2 & 85.9 & 1862 & 64.6
        & 100.0\% \\

        \addlinespace[2pt]
        \rowcolor{sectionbg}
        \multicolumn{8}{c}{\itshape Retain 192 Tokens ($\downarrow 66.7\%$)} \\

        ToME (ICLR'23)
        & 54.3 & 65.2 & 52.1 & 72.4 & 1563 & 60.5
        & 88.8\% \\

        FastV (ECCV'24)
        & 52.7 & 67.3 & 52.5 & 64.8 & 1612 & 61.2
        & 88.2\% \\

        HiRED (AAAI'25)
        & \textbf{58.7} & 68.4 & 47.4 & 82.8 & 1737 & 62.8
        & 93.6\% \\

        LLaVA-PruMerge (ICCV'25)
        & 54.3 & 67.9 & 54.3 & 71.3 & 1632 & 59.6
        & 90.3\% \\

        SparseVLM (ICML'25)
        & 57.6 & 69.1 & 56.1 & 83.6 & 1721 & 62.5
        & 95.9\% \\

        CaVIN (2026)
        & \textbf{58.7} & 69.0 & 55.8 & 84.4 & 1788 & 63.3
        & 97.0\% \\

        V2Drop (CVPR'26)
        & 58.5 & 69.3 & 55.6 & \textbf{85.1} & \textbf{1826} & 63.7
        & 97.6\% \\

        \rowcolor{oursbg}
        \textbf{E2S-Pruner (Ours)}
        & 58.6
        & \textbf{69.8}
        & \textbf{57.2}
        & 83.6
        & 1814
        & \textbf{64.4}
        & \textbf{98.0\%} \\

        \addlinespace[2pt]
        \rowcolor{sectionbg}
        \multicolumn{8}{c}{\itshape Retain 128 Tokens ($\downarrow 77.8\%$)} \\

        ToME (ICLR'23)
        & 52.4 & 59.6 & 49.1 & 62.8 & 1343 & 53.3
        & 80.4\% \\

        FastV (ECCV'24)
        & 49.6 & 60.2 & 50.6 & 59.6 & 1490 & 56.1
        & 81.7\% \\

        HiRED (AAAI'25)
        & 57.2 & 68.1 & 46.1 & 79.8 & 1710 & 61.5
        & 91.6\% \\

        LLaVA-PruMerge (ICCV'25)
        & 53.3 & 67.1 & 54.3 & 67.2 & 1554 & 58.1
        & 87.9\% \\

        SparseVLM (ICML'25)
        & 56.0 & 67.1 & 54.9 & 80.5 & 1696 & 60.0
        & 93.2\% \\

        CaVIN (2026)
        & 56.6 & 68.6 & 53.4 & 81.0 & 1702 & 61.2
        & 93.7\% \\

        V2Drop (CVPR'26)
        & 56.3 & 68.8 & 53.8 & 80.9 & 1712 & 61.8
        & 94.0\% \\

        \rowcolor{oursbg}
        \textbf{E2S-Pruner (Ours)}
        & \textbf{57.8}
        & \textbf{69.9}
        & \textbf{56.0}
        & \textbf{83.1}
        & \textbf{1779}
        & \textbf{63.5}
        & \textbf{96.8\%} \\

        \addlinespace[2pt]
        \rowcolor{sectionbg}
        \multicolumn{8}{c}{\itshape Retain 64 Tokens ($\downarrow 88.9\%$)} \\

        ToME (ICLR'23)
        & 48.6 & 50.0 & 45.3 & 52.5 & 1138 & 53.7
        & 72.3\% \\

        FastV (ECCV'24)
        & 46.1 & 51.1 & 47.8 & 48.0 & 1256 & 48.0
        & 71.3\% \\

        LLaVA-PruMerge (ICCV'25)
        & 51.9 & 68.1 & \textbf{54.0} & 65.3 & 1549 & 55.2
        & 86.5\% \\

        SparseVLM (ICML'25)
        & 52.7 & 62.2 & 51.8 & 75.1 & 1505 & 56.2
        & 86.5\% \\

        CaVIN (2026)
        & 51.2 & 68.5 & 51.2 & 75.4 & 1450 & 55.0
        & 86.7\% \\

        V2Drop (CVPR'26)
        & 50.5 & 68.9 & 51.8 & 75.1 & 1470 & 55.2
        & 86.9\% \\

        \rowcolor{oursbg}
        \textbf{E2S-Pruner (Ours)}
        & \textbf{55.5}
        & \textbf{69.4}
        & 49.5
        & \textbf{78.5}
        & \textbf{1656}
        & \textbf{57.3}
        & \textbf{90.6\%} \\

        \bottomrule
    \end{tabular}%
    }
\end{table*}

\begin{figure*}[!t]
\centering
\includegraphics[width=0.96\textwidth]{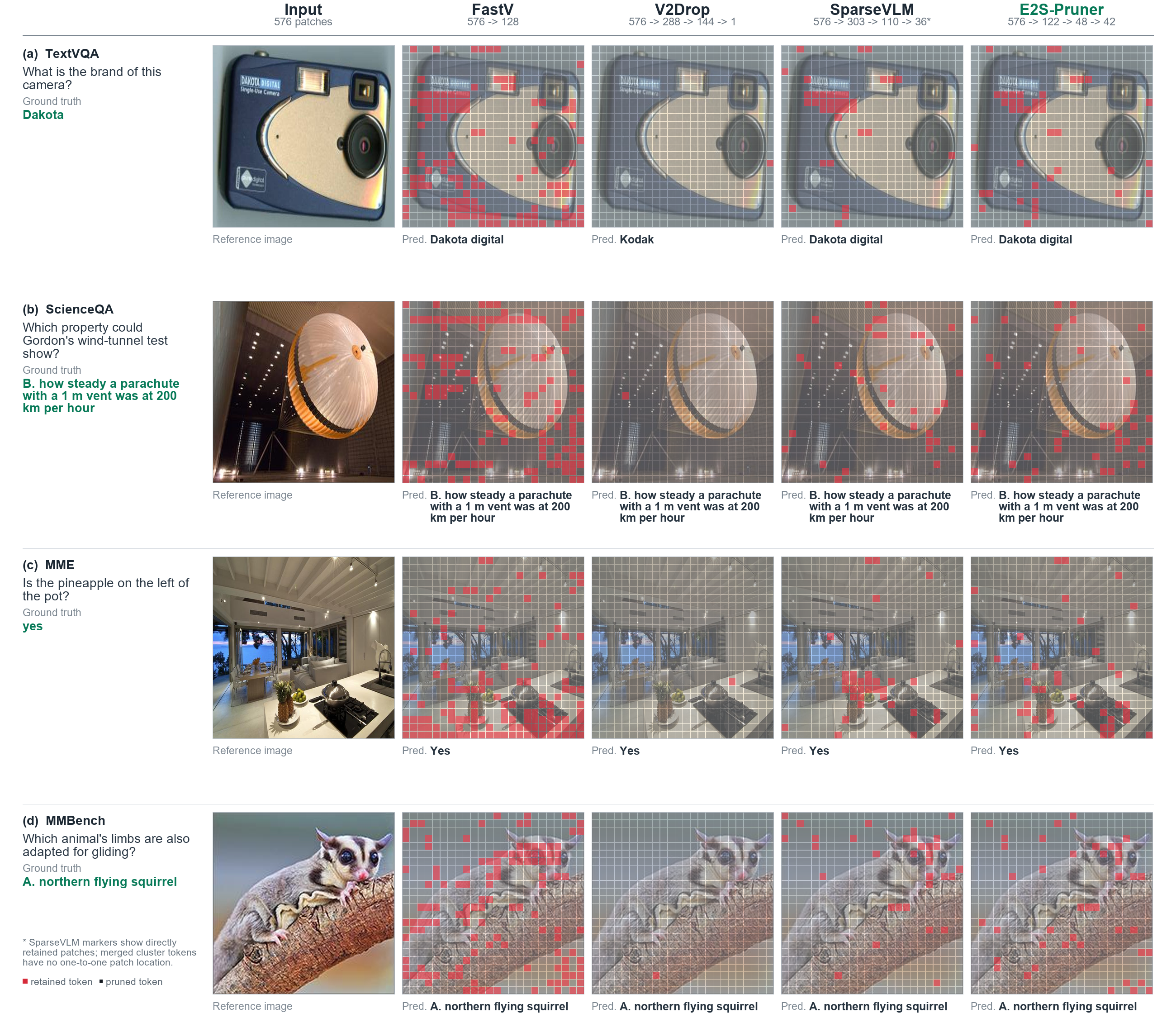}
\caption{Qualitative comparison of visual-token retention patterns and model outputs for representative test examples, illustrating how different pruning methods preserve task-relevant image regions.}
\label{Fig:intro}
\end{figure*}

\subsection{Ablation Experiment}
\begin{table}[t]
    \centering
    \scriptsize
    \setlength{\tabcolsep}{3.5pt}
    \renewcommand{\arraystretch}{1.12}
    \caption{Ablation study of the proposed E2S components.}
    \label{tab:e2s_ablation}
    \resizebox{\columnwidth}{!}{
    \begin{tabular}{lccccc}
        \toprule
        \textbf{Variant}
        & \textbf{SQA}
        & \textbf{TextVQA}
        & \textbf{MME}
        & \textbf{MMBench}
        & \textbf{Avg.} \\
        \midrule

        LLaVA-1.5-7B
        & 69.5 & 58.2 & 1862 & 64.6 & 100.0\% \\

        \midrule
        A1: Vanilla Attention Top-$K$
        & 64.9 & 55.8 & 1692 & 59.2 & 92.9\% \\

        A2: w/o Head Reliability
        & 68.0 & 55.8 & 1769 & 61.1 & 95.8\% \\

        A3: w/o Layer-wise Fusion
        & 68.9 & 55.8 & 1763 & 61.4 & 96.2\% \\

        A4: w/o Conflict Protection
        & 68.9 & 55.9 & \textbf{1781} & 62.3 & 96.8\% \\

        A5: w/o Spatial Novelty
        & 68.9 & 55.9 & 1774 & 61.2 & 96.3\% \\

        \midrule
        \rowcolor{oursbg}
        \textbf{Full E2S}
        & \textbf{69.9}
        & \textbf{56.0}
        & 1779
        & \textbf{63.5}
        & \textbf{97.7\%} \\

        \bottomrule
    \end{tabular}
    }
\end{table}

We evaluate each component of E2S-Pruner through ablation experiments at an average budget of 128 visual tokens. As shown in Table~\ref{tab:e2s_ablation}, all variants use identical models, datasets, prompts, evaluation protocols, and pruning locations; only the target component is removed or replaced. The complete method retains 97.7\% of aggregate performance while processing only approximately 22.2\% of the original visual-token budget.

\begin{itemize}
    \item \textbf{Complete evidence reasoning.} Replacing E2S-Pruner with conventional attention-based Top-$K$ selection (A1) reduces performance retention from 97.7\% to 92.9\%. The complete method improves SQA, TextVQA, MME, and MMBench by 5.0, 0.2, 87, and 4.3 points, respectively. Text-to-visual attention alone therefore does not identify critical evidence consistently, whereas explicit evidence modeling substantially improves token selection.
    \item \textbf{Multi-head reliability modeling.} Removing head-reliability estimation (A2) reduces retention to 95.8\%, 1.9 percentage points below the complete method; SQA, MME, and MMBench decrease by 1.9, 10, and 2.4 points. Evidence quality clearly differs across heads. Computing $\rho_{l,n}$ from evidence clarity and consistency reduces interference from ambiguous or contradictory responses.
    \item \textbf{Cross-layer evidence fusion.} Using only the current evidence layer at each pruning location (A3) reduces retention to 96.2\%. The complete method improves SQA, MME, and MMBench by 1.0, 16, and 2.1 points. Visual representations from different layers are complementary: combining shallow details, intermediate structures, and deep semantics reduces incidental single-layer bias and stabilizes token-importance estimates.
    \item \textbf{Conflict protection.} Removing conflict protection (A4) reduces retention to 96.8\%. The complete method improves SQA and MMBench by 1.0 and 1.2 points, indicating that tokens associated with inter-layer disagreement may still contain important information. Incorporating conflict $K$ through $q=1-(1-Pl)(1-K)$ reduces the risk of prematurely deleting such tokens. Although A4 attains a slightly higher MME score, its lower aggregate performance shows that conflict protection primarily improves cross-task stability.
    \item \textbf{Spatial novelty constraint.} Removing spatial novelty (A5) reduces retention to 96.3\%, with MME and MMBench decreasing by 5 and 2.3 points. Global Top-$K$ selection tends to concentrate tokens in a few high-response regions. Spatial novelty improves regional coverage and reduces the loss of information from other important areas.
\end{itemize}

\subsection{Analysis of Visual Token Retention Strategies}
\begin{table}[t]
    \centering
    \caption{Ablation study of different visual-token retention schedules.}
    \label{tab:token_schedule_ablation}

    \setlength{\tabcolsep}{3.5pt}
    \renewcommand{\arraystretch}{1.12}

    \resizebox{\columnwidth}{!}{%
    \begin{tabular}{lcccccc}
        \toprule
        \textbf{Schedule}
        & \textbf{SQA}
        & \textbf{TextVQA}
        & \textbf{MME}
        & \textbf{MMBench}
        & \textbf{Avg}
        & \textbf{Throughput} \\
        \midrule

        $576\rightarrow122\rightarrow48\rightarrow42$
        & 69.9 & 56.0 & 1779 & 63.5
        & 97.7\% & \textbf{10.37} \\

        $576\rightarrow132\rightarrow40\rightarrow32$
        & 69.9 & \textbf{55.9} & 1783 & 63.8
        & \textbf{97.8\%} & 10.57 \\

        $576\rightarrow142\rightarrow32\rightarrow22$
        & \textbf{70.1} & 55.6 & 1784 & 63.7
        & 97.7\% & 10.70 \\

        $576\rightarrow147\rightarrow38\rightarrow12$
        & 69.9 & 55.6 & 1782 & \textbf{63.9}
        & 97.7\% & 10.41 \\

        $576\rightarrow152\rightarrow44\rightarrow2$
        & 69.9 & 55.8 & \textbf{1785} & 63.7
        & 97.7\% & 10.71 \\

        \bottomrule
    \end{tabular}%
    }
\end{table}

To study how token allocation across layers affects performance, we evaluate five pruning schedules under the same average budget of 128 visual tokens. The schedules differ only in the token counts assigned to the three pruning stages; the model, datasets, pruning locations, and evaluation protocols remain unchanged.

As shown in Table~\ref{tab:token_schedule_ablation}, all five schedules retain 97.7\%--97.8\% of aggregate performance, indicating that E2S-Pruner is robust to the precise allocation of tokens across layers. The $576\rightarrow132\rightarrow40\rightarrow32$ schedule achieves the highest retention of 97.8\%, but its advantage is only 0.1 percentage points.

Different tasks favor different allocations. Retaining more visual tokens in deep layers, as in $576\rightarrow122\rightarrow48\rightarrow42$, gives the best TextVQA score of 56.0, suggesting that scene-text recognition benefits from continued visual participation in deep semantic reasoning. By contrast, allocating more tokens to intermediate layers increases MME from 1779 to 1785, indicating that richer intermediate visual information supports more complete semantic representations. SQA and MMBench vary only slightly and are comparatively insensitive to the allocation schedule.

Considering aggregate performance, TextVQA accuracy, and inference efficiency, we select
$576\rightarrow122\rightarrow48\rightarrow42$
as the default schedule. Although its aggregate retention is 0.1 percentage points below the maximum, it preserves more deep-layer visual information, performs better on text-intensive tasks, and maintains strong overall efficiency.

\subsection{Inference Efficiency Analysis}
\begin{table*}[t]
    \centering
    \small
    \setlength{\tabcolsep}{5.5pt}
    \renewcommand{\arraystretch}{1.16}
    \caption{Efficiency and performance comparison on LLaVA-1.5-7B, where percentages denote changes or performance retention relative to the dense baseline.}
    \label{tab:efficiency_comparison}
    \resizebox{\textwidth}{!}{
    \begin{tabular}{lccccc}
        \toprule
        \textbf{Method} &
        \makecell{\textbf{LLM Generation}$\downarrow$\\\textbf{Latency (s)}} &
        \makecell{\textbf{Total Latency}$\downarrow$\\\textbf{(s)}} &
        \makecell{\textbf{GPU Peak}$\downarrow$\\\textbf{Memory (MB)}} &
        \makecell{\textbf{Throughput}$\uparrow$\\\textbf{(item/s)}} &
        \textbf{Accuracy}$\uparrow$ \\
        \midrule

        LLaVA-1.5-7B
        & 676.11
        & 820.15
        & 15566
        & 5.28
        & 64.6 \\
        
        \rowcolor{sectionbg}
        \multicolumn{6}{c}{
            \itshape Avg. Retention 128 Tokens
            ($\downarrow 77.8\%$)
        } \\

        FastV (ECCV'24)
        & 334.87 \good{$\downarrow 50.5\%$}
        & 450.09 \good{$\downarrow 45.1\%$}
        & 15558 \good{$\downarrow 0.1\%$}
        & 9.73 \good{$\uparrow 1.84\times$}
        & 56.1 \good{86.8\%} \\

        SparseVLM (ICML'25)
        & 322.43 \good{$\downarrow 52.3\%$}
        & 435.72 \good{$\downarrow 46.9\%$}
        & 19229 \bad{$\uparrow 23.5\%$}
        & 10.05 \good{$\uparrow 1.90\times$}
        & 60.0 \good{92.9\%} \\

        V2Drop (CVPR'26)
        & 410.02 \good{$\downarrow 39.4\%$}
        & 605.69 \good{$\downarrow 26.1\%$}
        & \textbf{15046} \good{$\downarrow 3.3\%$}
        & 7.23 \good{$\uparrow 1.37\times$}
        & 61.8 \good{95.7\%} \\

        \rowcolor{oursbg}
        \textbf{E2S-Pruner (Ours)}
        & \textbf{312.48} \good{$\downarrow 53.8\%$}
        & \textbf{422.28} \good{$\downarrow 48.5\%$}
        & 18847 \bad{$\uparrow 21.1\%$}
        & \textbf{10.37} \good{$\uparrow 1.96\times$}
        & \textbf{63.5} \good{98.3\%} \\

        \midrule
        \rowcolor{sectionbg}
        \multicolumn{6}{c}{
            \itshape Avg. Retention 64 Tokens
            ($\downarrow 88.9\%$)
        } \\

        FastV (ECCV'24)
        & 310.86 \good{$\downarrow 54.0\%$}
        & 429.64 \good{$\downarrow 47.6\%$}
        & 15330 \good{$\downarrow 1.5\%$}
        & 10.19 \good{$\uparrow 1.93\times$}
        & 48.0 \good{74.3\%} \\

        SparseVLM (ICML'25)
        & 305.13 \good{$\downarrow 54.9\%$}
        & 412.34 \good{$\downarrow 49.7\%$}
        & 19122 \bad{$\uparrow 22.8\%$}
        & 10.62 \good{$\uparrow 2.01\times$}
        & 56.2 \good{87.0\%} \\

        V2Drop (CVPR'26)
        & 354.73 \good{$\downarrow 47.5\%$}
        & 535.86 \good{$\downarrow 34.7\%$}
        & \textbf{14978} \good{$\downarrow 3.8\%$}
        & 8.17 \good{$\uparrow 1.55\times$}
        & 55.2 \good{85.4\%} \\

        \rowcolor{oursbg}
        \textbf{E2S-Pruner (Ours)}
        & \textbf{286.11} \good{$\downarrow 57.7\%$}
        & \textbf{397.37} \good{$\downarrow 51.5\%$}
        & 18807 \bad{$\uparrow 20.8\%$}
        & \textbf{11.02} \good{$\uparrow 2.09\times$}
        & \textbf{57.3} \good{88.7\%} \\

        \bottomrule
    \end{tabular}
    }
\end{table*}

\begin{figure}[t]
\centering
\includegraphics[width=\columnwidth]{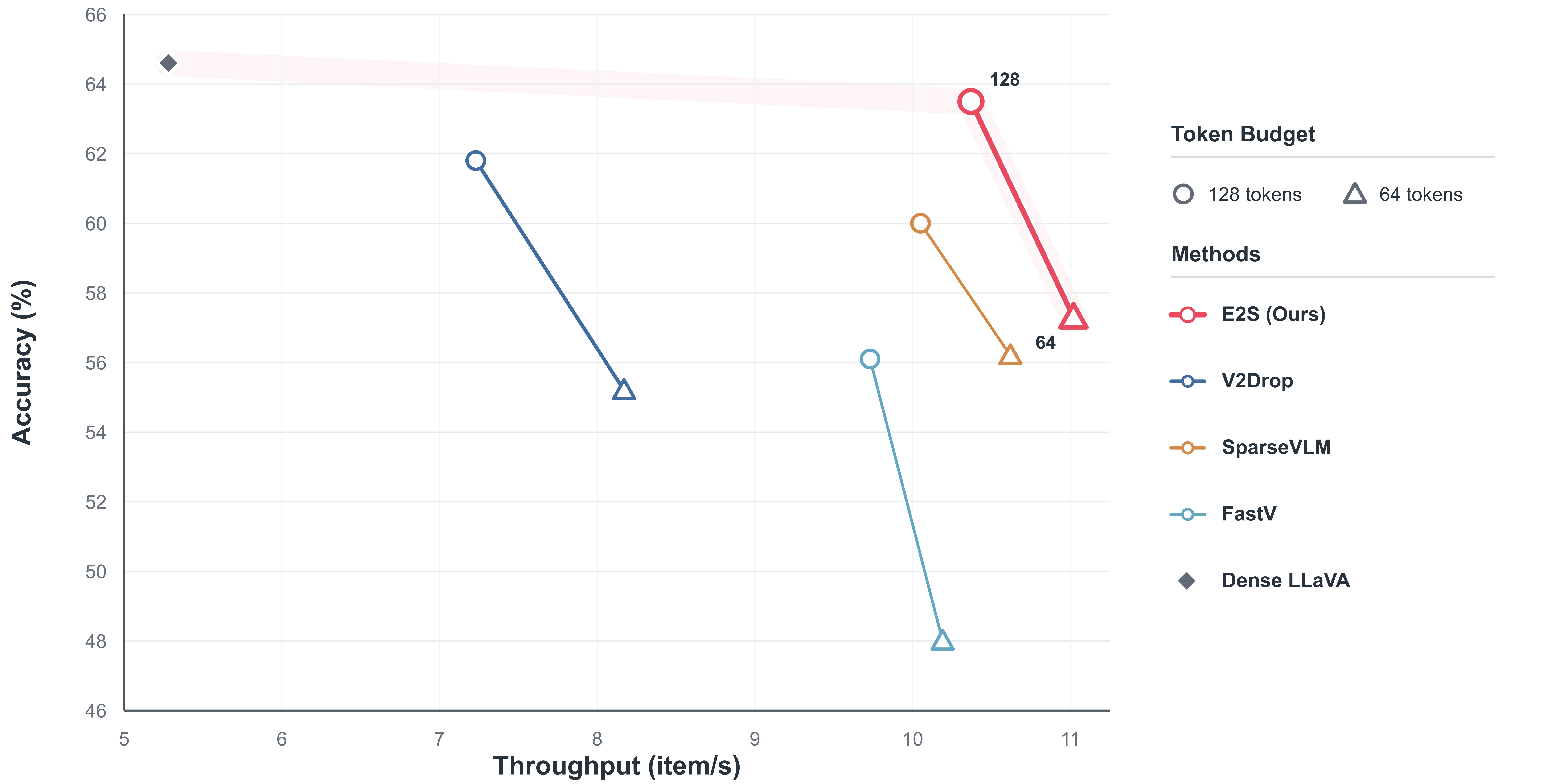}
\caption{Performance--throughput trade-off on LLaVA-1.5-7B under different average visual-token budgets, showing the practical acceleration achieved as the retained sequence becomes shorter.}
\label{Fig:throughput}
\end{figure}

To evaluate practical acceleration, we compare LLM generation latency, end-to-end latency, peak GPU memory, throughput, and task performance on MMBench. All relative changes use dense LLaVA-1.5-7B as the baseline. Table~\ref{tab:efficiency_comparison} reports the measurements, and Fig.~\ref{Fig:throughput} visualizes the performance--efficiency trade-off.

At an average budget of 128 visual tokens, E2S-Pruner reduces LLM generation latency from 676.11\,s to 312.48\,s (53.8\%) and end-to-end latency from 820.15\,s to 422.28\,s (48.5\%). Throughput increases from 5.28 to 10.37 items/s, corresponding to a $1.96\times$ speedup. Compared with SparseVLM, E2S-Pruner reduces end-to-end latency by a further 13.44\,s, improves throughput by approximately 3.2\%, and raises the task score from 60.0 to 63.5. Compared with V2Drop, it reduces end-to-end latency by 30.3\%, improves throughput by 43.4\%, and increases the task score by 1.7 points.

When the average budget is reduced to 64 visual tokens, E2S-Pruner decreases LLM generation and end-to-end latency by 57.7\% and 51.5\%, respectively. Throughput reaches 11.02 items/s, or $2.09\times$ that of the dense model, while retaining 88.7\% of its MMBench performance. Relative to SparseVLM, E2S-Pruner reduces end-to-end latency by 14.97\,s, improves throughput by approximately 3.8\%, and increases the task score by 1.1 points. Relative to V2Drop, the corresponding improvements are 25.8\%, 34.9\%, and 2.1 points.

As the budget decreases, E2S-Pruner consistently reduces generation latency, and throughput increases from 10.37 items/s at 128 tokens to 11.02 items/s at 64 tokens. The reduction in visual-token count therefore translates into practical acceleration rather than being offset by the additional evidence computation.

E2S-Pruner reaches peak GPU memory usages of 18,847 and 18,807\,MB under the two budgets, approximately 21\% above the dense baseline. This overhead arises mainly from intermediate states used for evidence modeling and cross-layer fusion. Although its peak memory is lower than that of SparseVLM, it remains higher than those of FastV and V2Drop. Reducing evidence caches and intermediate tensors is therefore an important optimization target. Overall, E2S-Pruner achieves the lowest end-to-end latency and highest throughput while preserving strong task performance, demonstrating its utility for efficient multimodal inference.

\subsection{Cross-Model Generalization Analysis}

To assess generalization across multimodal large language models, we apply E2S-Pruner to Qwen2-VL-7B and compare it with existing methods at two visual-token reduction ratios. As shown in Table~\ref{tab:qwen2vl_comparison}, E2S-Pruner retains 98.9\% and 95.4\% of dense-model aggregate performance at reduction ratios of 66.7\% and 77.8\%, respectively, achieving the highest average retention in both settings.

At a 66.7\% reduction ratio, E2S-Pruner retains 98.9\% of aggregate performance, exceeding FastV, DART, and V2Drop by 4.8, 2.1, and 1.3 percentage points. It achieves the best SQA and AI2D scores, 83.8 and 80.2, respectively. Although its POPE and MME results are slightly below the best individual scores, its superior aggregate result demonstrates more balanced preservation of visual question answering, hallucination assessment, and general perception capabilities.

At a 77.8\% reduction ratio, E2S-Pruner retains 95.4\% of aggregate performance, outperforming FastV, DART, and V2Drop by 4.5, 1.5, and 0.5 percentage points. Its SQA score reaches 82.9, 3.3 points above the second-best method, indicating that critical information for complex visual reasoning is preserved even at a low token budget. Although it does not achieve the best individual result on POPE, MME, or AI2D, its aggregate performance remains superior to all competing methods.

These results show that E2S-Pruner is not tied to a particular architecture or model family. Its evidence modeling and token-selection mechanisms transfer effectively to Qwen2-VL-7B and preserve stable aggregate performance across reduction ratios, confirming strong cross-model generalization.

\begin{table}[t]
    \centering
    \scriptsize
    \setlength{\tabcolsep}{3.2pt}
    \renewcommand{\arraystretch}{1.12}
    \caption{Comparison on Qwen2-VL-7B under different token reduction ratios.}
    \label{tab:qwen2vl_comparison}
    \resizebox{\columnwidth}{!}{
    \begin{tabular}{lccccc}
        \toprule
        \textbf{Method} &
        \textbf{SQA} &
        \textbf{POPE} &
        \textbf{MME} &
        \textbf{AI2D} &
        \textbf{Avg.} \\
        \midrule

        Qwen2-VL-7B
        & 84.7 & 86.1 & 2317 & 80.5 & 100.0\% \\

        \rowcolor{sectionbg}
        \multicolumn{6}{c}{\itshape Token Reduction ($\downarrow 66.7\%$)} \\

        FastV (ECCV'24)
        & 80.0 & 82.1 & 2130 & 76.1 & 94.1\% \\

        DART (EMNNLP'25)
        & 81.4 & 83.9 & \textbf{2245} & 78.0 & 96.8\% \\

        V2Drop (CVPR'26)
        & 81.6 & \textbf{87.2} & 2224 & 78.0 & 97.6\% \\

        \rowcolor{oursbg}
        \textbf{E2S-Pruner (Ours)}
        & \textbf{83.8} & 86.8 & 2228 & \textbf{80.2} & \textbf{98.9\%} \\

        \midrule

        \rowcolor{sectionbg}
        \multicolumn{6}{c}{\itshape Token Reduction ($\downarrow 77.8\%$)} \\

        FastV (ECCV'24)
        & 78.3 & 79.2 & 2031 & 73.8 & 90.9\% \\

        DART (EMNNLP'25)
        & 79.6 & 82.1 & 2175 & 74.4 & 93.9\% \\

        V2Drop (CVPR'26)
        & 78.9 & 85.1 & 2173 & \textbf{75.6} & 94.9\% \\

        \rowcolor{oursbg}
        \textbf{E2S-Pruner (Ours)}
        & \textbf{82.9} & \textbf{83.9} & \textbf{2156} & 74.9 & \textbf{95.4\%} \\

        \bottomrule
    \end{tabular}
    }
\end{table}

\section{Conclusion}

We presented E2S-Pruner, a progressive two-stage evidence-fusion method that reduces the inference cost caused by excessive visual tokens in VLMs. The method first estimates the reliability of evidence from different attention heads within each selected layer and then recursively fuses information across pruning layers using D--S evidence theory. Conflict protection reduces the risk of removing potentially important tokens associated with strong disagreement, while spatial novelty regularization promotes coverage of distinct image regions and prevents excessive concentration in locally salient areas.

On LLaVA-1.5-7B, E2S-Pruner retains 98.0\%, 96.8\%, and 90.6\% of aggregate performance at average budgets of 192, 128, and 64 visual tokens. Under the 128-token and 64-token settings, it reduces end-to-end latency by 48.5\% and 51.5\% and reaches $1.96\times$ and $2.09\times$ the baseline throughput. On Qwen2-VL-7B, it retains 98.9\% and 95.4\% of aggregate performance at two reduction ratios, demonstrating that its evidence modeling and spatial constraint generalize across model families.

E2S-Pruner nevertheless leaves room for improvement. Evidence fusion and spatial novelty computation require additional intermediate states and therefore increase peak GPU memory. Moreover, tasks such as TextVQA, which depend on fine-grained textual regions, remain vulnerable to information loss at extremely low token budgets. Future work will investigate lower-overhead evidence caching and parallel fusion, together with adaptive spatial protection for text and fine-grained objects, to improve deployment efficiency and robustness on resource-constrained devices.



{
    \small
    \bibliographystyle{ieeenat_fullname}
    \bibliography{main}
}




\end{document}